\documentclass[letterpaper, 10 pt, conference]{ieeeconf}  

\IEEEoverridecommandlockouts                              

\usepackage{xcolor}  
\usepackage{graphics} 
\usepackage{graphicx}
\usepackage[utf8]{inputenc}
\usepackage{amssymb}
\usepackage{amsmath} 
\usepackage{multirow}
\usepackage{booktabs}
\usepackage{hyperref}
\usepackage{cite}
\usepackage{array}
\usepackage{pifont}
\usepackage{bbding}
\hypersetup{hypertex=true,
colorlinks=true,
linkcolor=blue,
anchorcolor=blue,
citecolor=blue}

\usepackage{geometry}
\title{\LARGE \bf
Gaussian On-the-Fly Splatting: A Progressive Framework for Robust Near Real-Time 3DGS Optimization
}

\author{Yiwei Xu, Yifei Yu, Wentian Gan, Tengfei Wang, Zongqian Zhan, Hao Cheng and Xin Wang
\thanks{*This work was supported by the National Natural Science Foundation of China (No.42301507) and Natural Science Foundation of Hubei Province, China (No. 2022CFB727).}
\thanks{$^{1}$Yiwei Xu, Yifei Yu, Wentian Gan, Tengfei Wang, Zongqian Zhan, Xin Wang are with the School of Geodesy and Geomatics, Wuhan University, China PR. (Corresponding Author: Xin Wang, xwang@sgg.whu.edu.cn)
}%
\thanks{$^{2}$Hao Cheng is with the Department of Earth Observation Science at ITC Faculty Geo-Information Science and Earth Observation, University of Twente, the Netherlands.}%
}

\makeatletter

\newcommand{\Rmnum}[1]{\expandafter\@slowromancap\romannumeral #1@}
\makeatother

\begin{document}

\maketitle
\thispagestyle{empty}
\pagestyle{empty}

\begin{abstract}

3D Gaussian Splatting (3DGS) achieves high-fidelity rendering with fast real-time performance, but existing methods rely on offline training after full Structure-from-Motion (SfM) processing. 
In contrast, this work introduces Gaussian on-the-fly Splatting (abbreviated as On-the-Fly GS), a progressive framework enabling near real-time 3DGS optimization during image capture. 
As each image arrives, its pose and sparse points are updated via on-the-fly SfM\cite{43}$\footnote{See more at \href{ https://yifeiyu225.github.io/on-the-flySfMv2.github.io/}{ https://yifeiyu225.github.io/on-the-flySfMv2.github.io/}}$, and newly optimized Gaussians are immediately integrated into the 3DGS field.
To achieve this, we propose a progressive Local \& Semi-Global optimization to prioritize the new image and its neighbors by their corresponding overlapping relationship, allowing the new image and its overlapping images to get more training. 
To further stabilize training across previous and new images, an adaptive learning rate schedule balances the iterations and the learning rate.
Extensive experiments on multiple benchmarks show that our On-the-Fly GS reduces training time significantly, optimizing each new image in seconds with minimal rendering loss, offering one of the first practical steps toward rapid, progressive 3DGS reconstruction. 

\end{abstract}

\section{INTRODUCTION}

Novel View Synthesis (NVS) aims to generate realistic images of novel views based on a set of input images\cite{1}. Recently, Kerbl et al.\cite{3} introduced a novel 3D representation technique known as 3D Gaussian Splatting (3DGS), which employs a large number of 3D Gaussian kernels to represent a 3D scene. Subsequent research has made notable advancements in improving the 3DGS field in terms of training efficiency\cite{4,5,7,9,10,11}, rendering speed\cite{13,14}, and output quality\cite{15,16,17,18,19}. However, most existing approaches rely on an offline training solution, which means that 3DGS can only be optimized after all images are processed by COLMAP\cite{39}, preventing immediate synthesis of novel views upon completion of data acquisition\cite{20}. Some recent works\cite{20,21,44,45} have made progress towards real-time/near real-time training of 3DGS by integrating SLAM, such as GS-SLAM\cite{21}, RTG-SLAM\cite{20}. They take indoor video frames with inherent spatial and temporal continuity as input, which might degenerate in some practical applications\cite{41,43}. More detailed differences between SLAM-based 3DGS methods and our On-the-Fly GS are discussed in Sect.\ref{related work}.

\begin{figure}[tp]
    \centering
    \includegraphics[width=8.5cm]{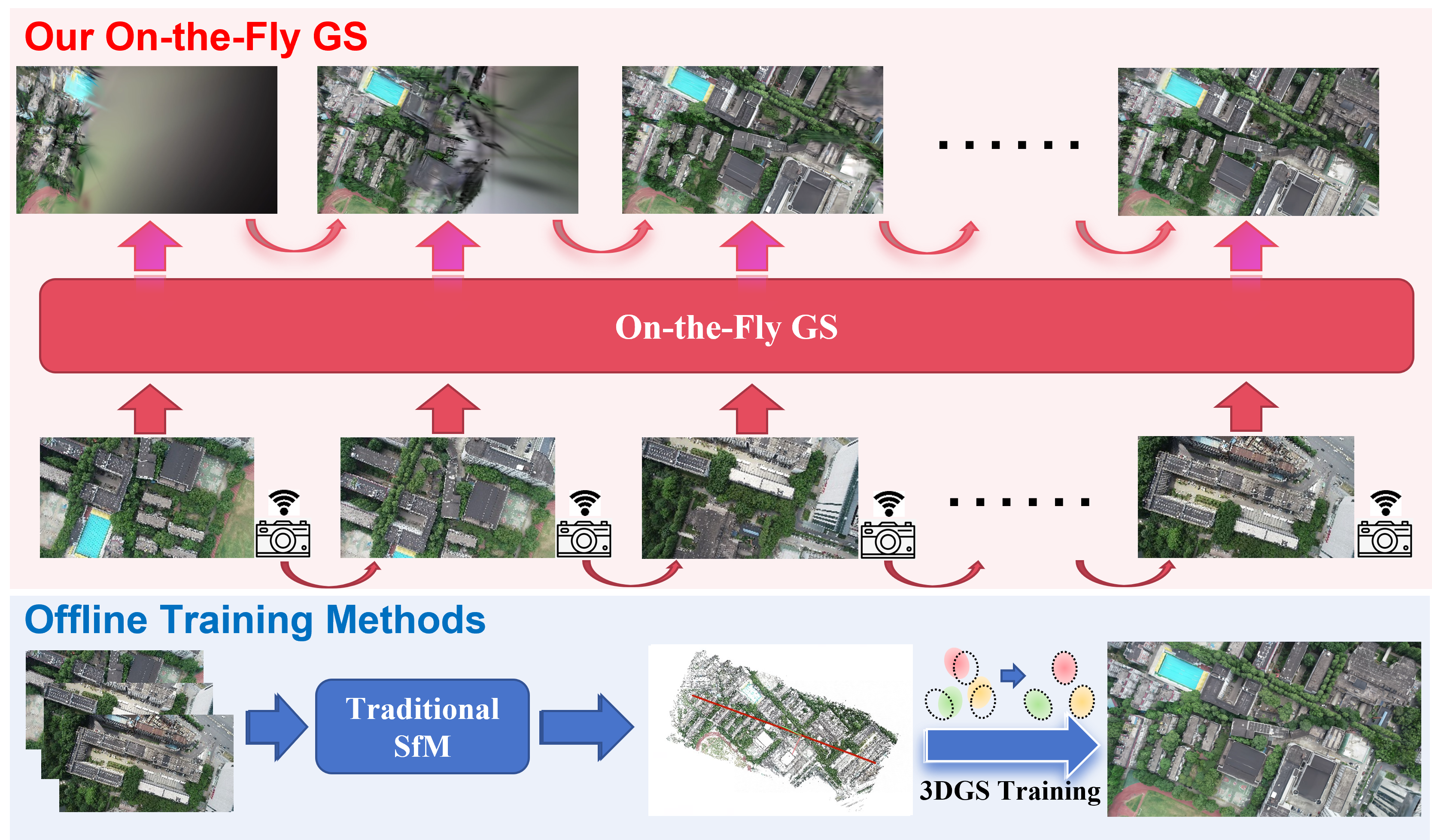}
    \caption{Our online On-the-Fly GS vs. Offline Training Method}
    \label{fig:1}
    \vspace{-0.6cm}
\end{figure}

To address these limitations, we propose On-the-Fly GS, a progressive framework for near real-time 3DGS field training that enables 3D Gaussian training during discrete image capturing. As shown in Fig. \ref{fig:1}, in our work, a new image can be acquired in an arbitrary manner. Its pose and relevant new point clouds are estimated online by the On-the-Fly SfM\cite{43}, which are then input into our On-the-Fly GS to optimize the corresponding local 3DGS field in near real time and update the entire 3DGS field.

In contrast to the original 3DGS, which focuses on reconstructing a static scenario using a fixed set of images and their corresponding poses, our On-the-Fly GS targets a dynamically expanding scenario, where images are progressively captured over time. This shift necessitates a different training strategy tailored for the incremental expansion of the 3DGS field. One key challenge in training dynamically extended scenes lies in the inconsistency of local quality within the 3DGS field. Images captured earlier undergo more training iterations, resulting in higher rendering quality, while newly acquired images, having undergone fewer training iterations, exhibit lower rendering quality. Consequently, the learning rate decay strategy of the original 3DGS based on complete 3DGS field quality is no longer applicable. Instead, On-the-Fly GS calculates the learning rate for each image based on the number of iterations it has undergone. Images with fewer training iterations are assigned higher learning rates, while images with more iterations receive lower learning rates. 

Furthermore, in the Local \& Semi-Global optimization, based on the overlapping relationships, each local connected image is assigned a weight that determines the number of training iterations it will undergo. This weight encourages prioritizing training the newly added images and their local overlapping images, thereby accelerating the rendering quality in regions captured more recently. To further enhance training efficiency and near real-time performance for each image, two strategies have been integrated into On-the-Fly GS: inspired by Adr-Gaussian\cite{5}, a load-balancing strategy is incorporated to mitigate the over-accumulation of 3D Gaussians, while simultaneously ensuring fast training of newly added regions; Drawing from insights from Taming-3DGS\cite{9}, we re-implement the back-propagation by transitioning from pixel-based parallelism to splatting-based parallel computation. In summary, this work makes the following contributions:
\begin{itemize}
    \item We propose On-the-Fly GS, a novel progressive solution for near real-time 3DGS field optimization that enables robust online training during image acquisition.
    \item To achieve robust and near real-time progressive 3DGS training, a Local \& Semi-Global optimization strategy is designed, and self-adaptive methods are presented to determine the training iteration number of each image and to update the learning rate during each iteration.
    \item Our On-the-Fly GS reduces training time significantly on several benchmark datasets, offering one of the first practical steps toward rapid 3DGS optimization during image capturing.
\end{itemize}

\section{RELATED WORKS}\label{related work}

\noindent\textbf{Novel View Synthesis (NVS).} Traditional photogrammetry method takes unposed images as input data, generates a mesh model of the current scene through a complex pipeline, and renders images from arbitrary viewpoints using back-projection\cite{2}. In recent years, the advent of Neural Radiance Fields (NeRF)\cite{28} and 3D Gaussian Splatting (3DGS)\cite{3} has introduced novel paradigms for NVS. NeRF, based on volumetric rendering, leverages a multi-layer perceptron (MLP) to implicitly represent a 3D scene. Although NeRF has demonstrated strong rendering capabilities, it still requires substantial computational resources when facing significant challenges in complex, unbounded outdoor scenes, where achieving high-quality, high-resolution rendering. To address these limitations, Kerbl et al.\cite{3} introduced 3DGS to explicitly represent 3D scenes. 3DGS offers faster training and rendering speeds for novel views while maintaining acceptable rendering quality, which has led to widespread attention since its introduction.

\vspace{4pt}
\noindent\textbf{3DGS fast training.} Although 3DGS delivers high-quality and real-time rendering, its training process remains computationally intensive, making it difficult to directly meet the demands of near real-time or real-time applications. Consequently, extensive research has been dedicated to accelerating the 3DGS optimization.


Several studies have sought to expedite 3DGS training by reducing the number of Gaussians in the scene representation. Panagiotis et al.\cite{4} proposed an efficient and resolution-aware Gaussian pruning strategy along with an adaptive adjustment method for the degree of spherical harmonic coefficients. It reduces the number of Gaussians by more than 50\% and saves memory. EfficientGS\cite{7} and LightGaussian\cite{8} evaluate the contribution of each Gaussian, only retaining the critical Gaussians and removing the redundant ones. Adr-Gaussian\cite{5} introduced a load balancing strategy to optimize the distribution of Gaussians and prevent their excessive accumulation in specific regions while also avoiding too few Gaussians in other areas. Since 3DGS training requires optimizing the parameters of a large number of Gaussians, reducing 3D Gaussians can significantly enhance training speed and decrease memory overhead.



Unlike methods that focus on reducing the number of Gaussians, some studies also focused on refining training strategies to improve computational efficiency. Taming 3DGS\cite{9} implements a Gaussian-based parallel strategy for gradient accumulation and optimizes the computation of spherical harmonics and differentiable loss functions, thereby enhancing back-propagation efficiency. Facing the challenge of large-scale scenes and high-resolution image data, RetinaGS\cite{10} and DOGS\cite{11} propose a parallel training method that divides the whole scene into several overlapping blocks, enabling parallel and accelerating training for large-scale scenes.

\vspace{4pt}
\noindent\textbf{3DGS real-time training.}  Recent studies have explored
 real-time training of 3DGS. Before 3DGS, through integration with SLAM, NICE-SLAMM\cite{44} and Vox-Fusion\cite{45} successfully achieved real-time NeRF training. More recently, 3DGS has also achieved real-time training via similar integration with SLAM. GS-SLAM\cite{21} predicts and continuously refines the poses of keyframes. Based on depth priors, it adds a certain number of Gaussians to the 3DGS field, only focusing the optimization primarily on newly added and recently added keyframes. RTG-SLAM\cite{20} also leverages depth to support 3DGS training and further improves the strategies for both Gaussian addition and optimization. Compared to GS-SLAM, it achieves significantly better training efficiency and novel view rendering quality. Subsequently, numerous efforts have been made to further improve SLAM-based 3DGS methods. Deng et al.\cite{51} propose a sliding window-based masking strategy to reduce redundant Gaussians, thereby improving training efficiency and reducing memory consumption. GSFusion\cite{50} incorporates each Gaussian into a volumetric mapping system to leverage geometric information, significantly accelerating the training process while maintaining high rendering quality. However, their reliance on depth sensors would be limited when dealing with large-scale outdoor scenes or UAV images. Although some existing methods take RGB-only data as input (such as GS-SLAM\cite{21}, WildGS-SLAM\cite{54}, DROID-Splat\cite{55}, On-the-Fly NVS\cite{56}), they still struggle to handle discrete images that are not spatially and temporally continuous.
 
 In contrast, our On-the-Fly GS, which can deal with images captured in an arbitrary way without spatiotemporal continuity, employs a progressive framework for near real-time 3DGS training. It only requires that the newly fly-in image has a certain degree of overlap with all previously registered images, and does not rely on depth priors for 3DGS field expansion. It addresses the limitations of existing SLAM-based 3DGS methods and extends the applicability of 3DGS training to more general use cases.



\section{PRELIMINARIES}
In this section, two key preliminaries of 3DGS\cite{3} and On-the-fly SfM\cite{43} are introduced.

\begin{figure*}[t!]
    \centering
    \includegraphics[trim={0.65cm 0 0 0},clip, width=0.997\textwidth]{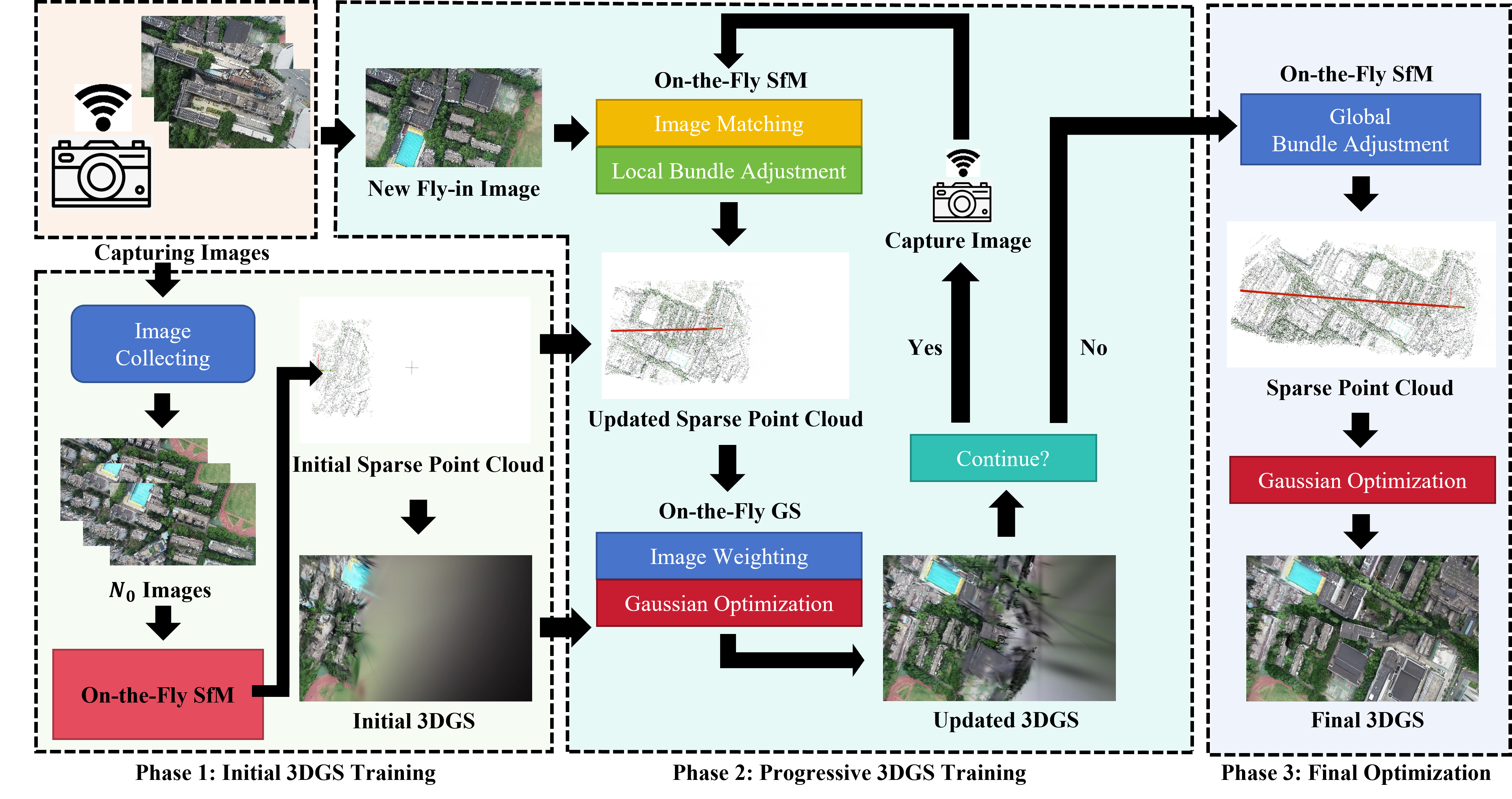}
    \caption{Workflow of the proposed On-the-Fly GS. We first initialize the 3DGS field. Then, for each new fly-in image, we update the current 3DGS field and perform the optimization on it. Finally, after a final global BA of previous SfM results, we further refine the 3DGS field to enhance rendering quality.}
    \label{fig:2}
     \vspace{-0.6cm}
\end{figure*}

\noindent\textbf{3D Gaussian Splatting.} 3DGS employs a large number of Gaussians to represent a real 3D scene\cite{3}. Given the rotation matrix $R$ and the scaling matrix $S$, the geometric properties of each Gaussian are defined by a covariance matrix $\Sigma=RSS^{T}R^{T}$ centered at point $X_0$:


$$
G(X)=e^{-\frac{1}{2}\left(X-X_{0}\right)^{T} \boldsymbol{\Sigma}^{-1}\left(X-X_{0}\right)} \eqno{(1)}
$$



Furthermore, 3DGS uses spherical harmonics (SH) and opacity $\sigma$ to reveal the material properties. Given the splatting opacities $\alpha_i$ estimated based on Gaussian-to-Pixel distance and color $c_i$ calculated from SH of $N$ Gaussians associated with the specific pixel, 3DGS calculates the color of the pixel $c_p$ by alpha blending, as follows:




$$
c_\text{pixel}=\sum_{i=1}^{N}\left ( c_{i}a_{i}\prod_{j=1}^{i-1}\left ( 1-a_{j}   \right )    \right ) \eqno{(2)}
$$


\noindent\textbf{On-the-Fly SfM.} Traditional SfM frameworks (e.g., COLMAP\cite{39}) are incapable of processing images in an online manner. To overcome this limitation, we rely on the On-the-Fly SfM\cite{43} for our subsequent implementation. On-the-Fly SfM employs a Hierarchical Weighted Local Bundle Adjustment (BA), in which only neighboring connected and registered images are incorporated into BA after acquiring a new image. Unlike COLMAP’s BA, which incurs a substantial computational cost as the number of images increases, Hierarchical Weighted Local BA significantly reduces time complexity, ensuring efficient processing. Additionally, On-the-Fly SfM can achieve online feature matching via global feature retrieval on HNSW (Hierarchical Navigable Small World)\cite{48}, enabling efficient dynamic updates for the matching matrix. On-the-fly SfM allows for near real-time pose estimation and sparse point cloud generation immediately after capturing a new image, ensuring a continuous and timely input for our approach.


\section{METHODOLOGY}

\subsection{Overview of on-the-fly GS}

To achieve near real-time 3DGS field optimization, we propose a progressive training framework. As illustrated in Fig. \ref{fig:2}, our training process is structured into three distinct phases:


\noindent\textbf{Phase 1: Initial 3DGS Training.} Training the 3DGS field with a limited number of images risks overfitting to the initial few images (e.g., 2 or 3 images)\cite{34}. To mitigate this, we begin by training an initial 3DGS field after On-the-Fly SfM has estimated the poses and generated the sparse point cloud for $N_0$ initial images. This initial 3DGS field serves as the foundation for subsequent progressive training. In this phase, we adopt the original 3DGS training method, but a stricter limit was imposed on the number of training iterations to prevent overfitting to these initial images.

\noindent\textbf{Phase 2: Progressive 3DGS Training.} As a new image is acquired, On-the-Fly SfM continuously updates the camera poses of both the newly acquired and previously registered images, while refining the sparse point cloud. Based on the updated sparse point cloud, some new Gaussians are appended to the previous 3DGS field. The updated 3DGS field will then be optimized. During the optimization, the new fly-in image and its neighboring images are prioritized for training to accelerate the optimization of newly expanded regions. Meanwhile, some of the other images are randomly selected as training targets to avoid degrading the entire scene, including the very previous 3DGS field.


\noindent\textbf{Phase 3: Final Optimization.} Once all images have been acquired and the progressive training is complete, On-the-Fly SfM further performs global BA on the estimated poses. Subsequently, a rapid additional optimization is applied to the current 3DGS field, enabling a rapid improvement in overall rendering quality within a short period of time.


\subsection{3D Gaussian field Update with newly fly-in images}

For the newly fly-in image, On-the-Fly SfM estimates the pose of the new image and optimizes the pose of its neighboring connected images via an improved local BA\cite{43}, simultaneously updating the sparse point cloud. Before updating the previously trained 3DGS field, a search tree is constructed for the existing sparse point cloud. The sparse points from the new image are then queried against this search tree using a predefined threshold. Points in the new image whose shortest distance from the existing points exceeds the threshold are classified as newly added points. These newly added points are then initialized as newly added Gaussians and directly incorporated into the current 3DGS field, facilitating the 3DGS field training on the newly expanded region.

\subsection{Self-adaptive Training Iteration Allocation via Hierarchical Image Weighting}

\begin{figure}[htp]
    \centering
    \includegraphics[width=8.7cm]{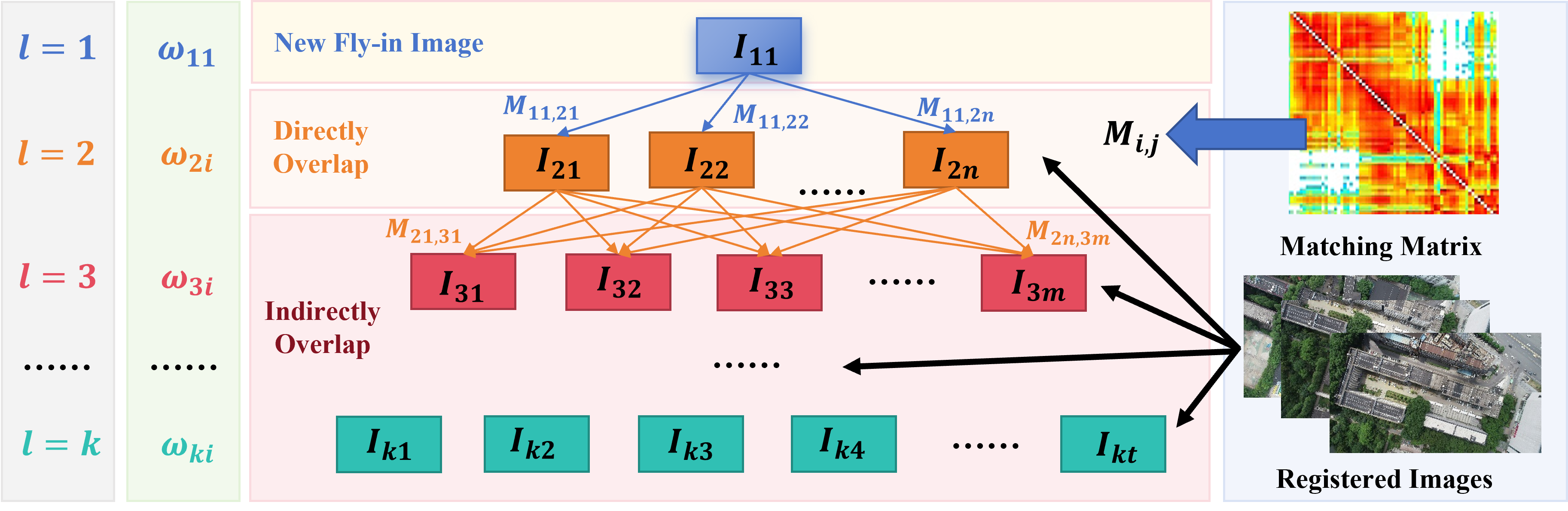}
    \caption{Hierarchical Image Weighting. The already solved images are clustered into hierarchical layers using their overlap relationships, assigning higher weights to those that exhibit closer relationship with the new image.}
    \label{fig:3}
    \vspace{-0.3cm}
\end{figure}

\noindent\textbf{Hierarchical Image Weighting.} Newly added images often introduce novel, unexplored regions or correspond to areas where the current 3DGS field exhibits poor rendering quality. To ensure that more training is allocated to the new image and its neighboring connected images during optimization, we propose an image hierarchical weighting strategy, which can be used to adaptively allocate training iterations for each image. The weight is determined by the overlap relationship between the newly fly-in image and the already registered images, as depicted in Fig. \ref{fig:3}.

Assuming that there are $n-1$ registered images, when the image $n$ ($I_{11}$, $l=1$) is newly added, the weight of the newly added image is fixed as $\omega_{11}=1$. The weights of the remaining registered images are calculated based on the match matrix online, estimated by On-the-Fly SfM. In the match matrix, the overlapping degree of any two images is calculated based on the number of matched feature points shared between them. If the overlapping degree of any two images is $M_{i,j}$, for images $I_{2i}$ that have a direct overlapping relationship ($M_{i,n}\neq0,l=2$) with the newly fly-in image, their weights $\omega_{2i}$ are calculated using the following equation:
$$
   \omega _{2i}=\omega _{11}\times   M_{i,n}= M_{i,n}. \eqno{(3)}
$$
For images $I_{li}$ that have an indirect overlapping relationship ($M_{i,n}=0,l\geq3$) with the newly added image, their weights are determined by their overlapping relationship with images in the previous layer and the weights of those images. If the layer $k-1$ contains $N_{k-1}$ images, the weight of the image $j$ in the layer $k-1$ is $\omega_{(k-1)j}$, the weight of the corresponding image $i$ in layer $k$ will be calculated as follows:

$$
\omega _{ki}=\frac{ {\textstyle \sum_{j=1}^{N_{k-1}}}\omega_{\left ( k-1 \right )j  }\times M_{\left ( k-1 \right )j,ki }    }{N_{k-1}}    \eqno{(4)}
$$

\noindent\textbf{Local \& Semi-Global optimization.} After the addition of any new fly-in image, we perform $T_I$ training iterations to optimize the current 3DGS field. The training process is structured into two parts of local and semi-global ones, with each performing $T_I$/2 training iterations. \textbf{1) Local optimization:} The top $N_m$ images with the highest weights will be selected as key training images. For one key training image $i$ with corresponding weight $\omega_k ( k=1,2,...,N_m$), it is supposed to be trained for $T_i^{m}$ iterations, which is determined by Eq 5. \textbf{2) Semi-Global optimization:} For the other images with low weights, the remaining half training iterations are evenly distributed among these images. This strategy ensures that more training is performed on the new fly-in image and its neighbor images, while avoiding forgetting the 3DGS field of the very previous scene.



$$
T_i^{m} = \frac{1 - e^{-\omega_{i}}  }{ 2\sum_{j=1}^{N_m}\left (1 - e^{-\omega_{j}}\right )  }\times T_I   \eqno{(5)}
$$




\subsection{Adaptively Learning Rate Updating}

The progressive training in On-the-Fly GS may result in inconsistency in local quality within the 3DGS field. To address this issue, we dynamically update the learning rate based on the rendering quality of each image, instead of updating the learning rate based on the overall quality of the 3DGS field (this was applied in the original 3DGS). Specifically, for image $j$, if it has been trained for $\text{T}_j$ iterations, with the initial learning rate $\eta_j^{0}$ and the final learning rate $\eta_j^{f}$, the current learning rate $\eta_j^c$ can be calculated as follows:

$$
\eta_{j}^c=e^{ \ln\left ( \eta_j^{0} \right )\times \left ( 1-t_{i}  \right ) +  \ln\left ( \eta_j^{f} \right ) \times t_{i}    }, \eqno{(6)}
$$

$$
t_{i}=\left\{\begin{matrix}1,
  & \text{T}_{j} > \text{T}_{a} \\ \frac{\text{T}_{j} }{\text{T}_{a}}, 
  & \text{T}_{j} \le \text{T}_{a} 
\end{matrix}\right., \eqno{(7)}
$$
where $\text{T}_{a}$ is a predefined parameter that represents the number of training iterations required for an image to achieve satisfactory rendering quality. However, determining the learning rate based on the already trained iterations of a single image is not rigorous. This is because other images overlapping with image $j$ might already achieve satisfactory rendering quality. Thus, when determining the current learning rate, it makes sense to also consider the overlapping images among the registered images. If the set of overlapping images $\mathcal{I}$ for the current image $j$ contains $N_{\mathcal{I}}$ images (i.e., for any $i\in \mathcal{I}, M_{i,j}>0$). The learning rate $\eta_j^{\mathcal{I}}$ for image $j$ can be calculated as follows:

$$
\eta_{j}^{\mathcal{I}} =\frac{\eta_{j}^c +  {\textstyle \sum_{i=1}^{N_{\mathcal{I}}}} M_{i,j} \times \eta_{i}^c    }{N_{\mathcal{I}}+1}.  \eqno{(8)}
$$

\begin{figure}[htp]
    \centering
    \includegraphics[trim={0 0 0 0},clip, width=\linewidth]{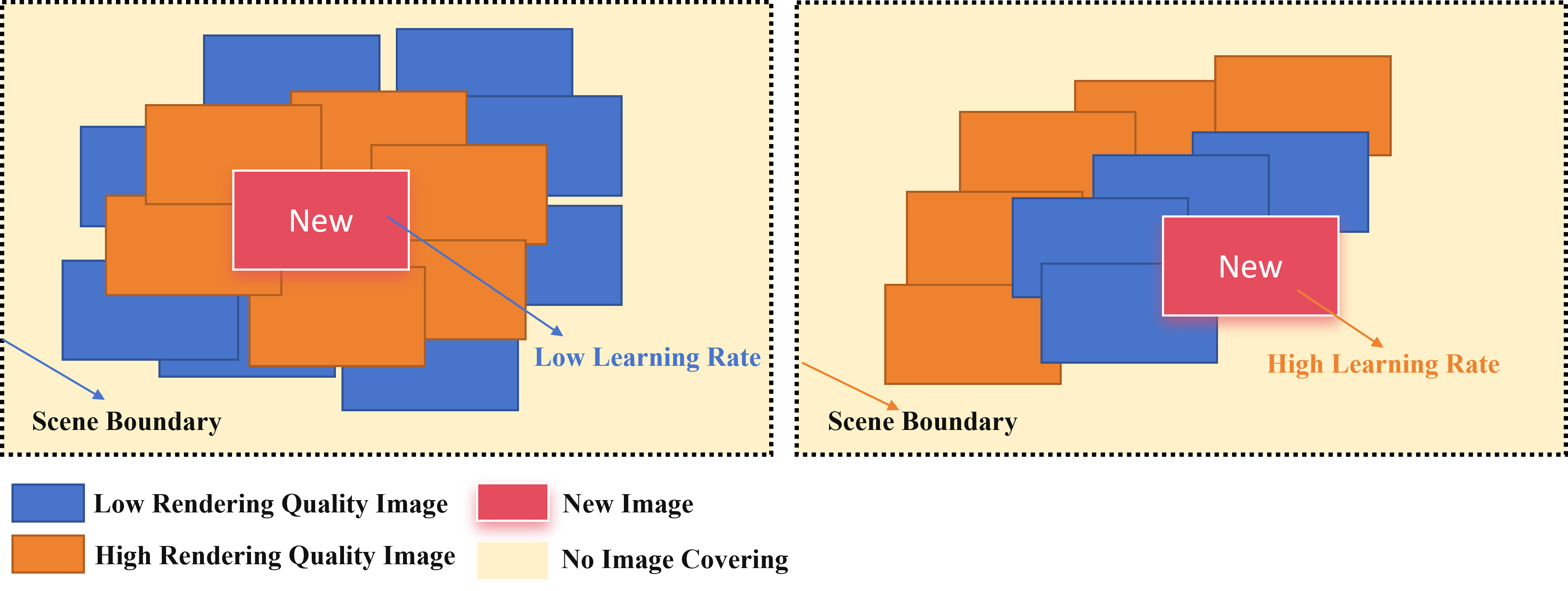}
    \caption{Adaptively Learning Rate Updating}
    \label{fig:4}
    \vspace{-0.4cm}
\end{figure}

The proposed learning rate updating method considers the rendering quality of the 3DGS field within the local scene. If the local scene corresponding to the current image $j$ has achieved high rendering quality, the learning rate $\eta_{j}^{\mathcal{I}}$ is supposed to be lower. This is because multiple images overlapping with image $j$ have been trained for more iterations (even if image $j$ is newly added), as shown in Fig. \ref{fig:4} (left). If the local scene corresponding to the image $j$ performs poor rendering quality, which indicates that both image $j$ and its neighboring connected images are only trained for fewer iterations, the learning rate $\eta_{j}^{\mathcal{I}}$ should be higher, as shown in Fig. \ref{fig:4} (right).

\subsection{Load Balancing and 3DGS Runtime Optimization}

Some local areas may undergo excessive Gaussian accumulation due to being visible across multiple images, while newly introduced scenes may suffer from insufficient Gaussian kernels. To address this uneven distribution of Gaussians, inspired by Adr-Gaussian\cite{5}, we employ a load-balancing training. For the current training image $i$, we record the number of Gaussians $\text{G}_p$ associated with each pixel $p$ in image $i$. Using the standard deviation of the Gaussian distribution within the visible range of the image, we define a Load-Balancing-Loss $L_\text{load}$ as follows:

$$
L_\text{load}=\text{std}\left ( \text{G}_{p}  \right ), p\in i .  \eqno{(9)}
$$
With L1-Loss and SSIM-Loss, the total loss is as follows:
$$
\text{L}_\text{total}=\lambda _{1}L_{1}+  \lambda_\text{ssim}L_\text{ssim}+\lambda_\text{load}L_\text{load}. \eqno{(10)}
$$

Subsequent experiments demonstrate that the load-balancing strategy not only prevents the abnormal accumulation of Gaussians in some specific regions but also effectively reduces the total number of Gaussians in the 3DGS field, thereby accelerating the overall training process. Furthermore, it facilitates the densification of Gaussians in newly added regions, promoting faster 3DGS training.

To further accelerate the training of the 3DGS field, we integrate the strategy proposed by Taming 3DGS\cite{9}. Taming 3DGS identifies that the per-pixel gradient accumulation method used in the original 3DGS leads to multiple threads contending for accessing the same locations and thus serialized atomic operations. To address this, we implemented the corresponding Splatting-based parallelism in the gradient accumulation method. In this implementation, gradients are accumulated directly for each Gaussian, significantly reducing training time by eliminating contention and enhancing parallel processing efficiency.

\section{EXPERIMENT}

\subsection{Experimental Settings}

\noindent\textbf{Datasets and Metrics}. To validate the efficacy of our On-the-Fly GS, we utilizes several datasets from the 3DGS (RGB)\cite{3}, Mip-NeRF 360 (RGB)\cite{42}, On-the-Fly SfM (RGB)\cite{41,43}, and Replica (RGB-D)\cite{49}. These contain three indoor scenes, three small-scale outdoor scenes, three UAV datasets and eight VSLAM datasets, which are dynamically processed as simulated sequential inputs according to the original stored order. Based on these datasets, the corresponding performance is evaluated from three aspects: Training efficiency, Rendering quality, and Model size. All experiments are conducted under the same hardware configuration (Single NVIDIA RTX 4090 GPU).


\begin{table*}[]
\centering
\caption{Comparison on RGB datasets. Best and second-best are highlighted in bold and underlined.}
\vspace{-0.2cm}
\label{Comparison0}

\begin{tabular*}{0.98\linewidth}{cc|ccc|ccc|ccc|c}
\toprule
\multicolumn{2}{c}{Type}                                & \multicolumn{3}{c}{Indoor}                         & \multicolumn{3}{c}{Outdoor}                                       & \multicolumn{3}{c}{UAV}                            &                       \\ \midrule
\multicolumn{2}{c|}{Datasets}                            & skfx             & kitchen        & counter        & train          & Temple         & JYYL           & WHU            & npu             & phantom3        & \multirow{3}{*}{Avg.} \\
\multicolumn{2}{c|}{Image Number}                        & 60               & 279            & 240            & 301            & 302            & 390            & 242            & 300             & 399             &                       \\ 
\multicolumn{2}{c|}{Pose Estimation (s)}                 & 95.48            & 1037.27        & 615.67         & 813.98         & 1068.51        & 1373.87        & 820.06         & 808.21          & 1183.37         &                       \\ \midrule
\multirow{5}{*}{\begin{tabular}[c]{@{}c@{}}Adr-Gaussian\\ (7k)\end{tabular}} & PSNR (dB)          & 31.07            & 28.45          & 26.81          & 19.94          & 18.68          & 19.85          & 24.99          & 26.85           & 26.83           & 24.83\\
                                   & SSIM               & 0.87             & 0.89           & 0.86           & 0.69           & 0.67           & 0.69           & 0.82           & 0.81            & 0.83            & 0.79                  \\
                                   & LPIPS              & 0.28             & 0.15           & 0.24           & 0.36           & 0.57           & 0.47           & 0.17           & 0.29            & 0.26            & 0.31                  \\
                                   & Model Size (MB)    & \underline{323.4}& 192.5          & \underline{104.5}& \underline{63.2}& \textbf{21.3}& \underline{92.8}& 839.2   & 895.5           & \underline{809.1}& 371.3                \\
                                   & Ad-Opt Time (s)    & 176.36           & 245.96         & 207.96         & 109.36         & 208.59         & 280.58         & 268.09         & 268.98          & 279.15          & 227.23                \\ \midrule
\multirow{5}{*}{\begin{tabular}[c]{@{}c@{}}Taming-3DGS\\ (7k)\end{tabular}}& PSNR (dB)& \textbf{32.43}& \underline{29.41}& \underline{27.52}& \underline{20.57}& \underline{19.39}& \underline{20.16}& \underline{25.33}& \underline{27.25}& \underline{27.16}& \underline{25.47}\\
                                   & SSIM               & 0.89             & \underline{0.90}& \underline{0.88}& \underline{0.72}& \underline{0.69}& \underline{0.71}& \underline{0.84}& \underline{0.83}& \underline{0.84}& \underline{0.81}\\
                                   & LPIPS              & 0.22             & \underline{0.13}& \underline{0.21}& \underline{0.33}& \underline{0.54}& \underline{0.44}& \underline{0.14}& \underline{0.26}& \underline{0.23}& \underline{0.28}\\
                                   & Model Size (MB)    & 652.7            & 386.9          & 217.6          & 149.7          & \underline{63.1}           & 241.8          & 1648.4         & 1784.8          & 1596.7          & 749.1                 \\
                                   & Ad-Opt Time (s)    & \underline{113.65}     & 155.67         & 123.87         & 75.34          & 88.39          & 103.47         & 245.62         & 236.58          & 240.92          & 153.72           \\ \midrule
\multirow{5}{*}{\begin{tabular}[c]{@{}c@{}}Taming-3DGS \\ (30k)\end{tabular}} & PSNR (dB)       & 37.84  & 33.01   & 30.16   & 25.35  & 22.58  & 23.93  & \textbackslash{} & \textbackslash{} & \textbackslash{} & 28.81                 \\
                                                                              & SSIM            & 0.95   & 0.94    & 0.92    & 0.86   & 0.77   & 0.82   & \textbackslash{} & \textbackslash{} & \textbackslash{} & 0.88                  \\
                                                                              & LPIPS           & 0.09   & 0.08    & 0.15    & 0.15   & 0.37   & 0.25   & \textbackslash{} & \textbackslash{} & \textbackslash{} & 0.18                  \\
                                                                              & Model Size (MB) & 950.4  & 417.7   & 250.2   & 233.1  & 122.6  & 504.8  & \textbackslash{} & \textbackslash{} & \textbackslash{} & 413.1                 \\
                                                                              & Ad-opt Time (s) & 950.34 & 953.09  & 475.13  & 350.17 & 343.09 & 612.72 & \textbackslash{} & \textbackslash{} & \textbackslash{} & 614.09                \\ \midrule
\multirow{6}{*}{Ours}              & PSNR (dB)          & \underline{31.93}& \textbf{30.39} & \textbf{28.97} & \textbf{22.82} & \textbf{21.72} & \textbf{22.71} & \textbf{27.93} & \textbf{33.29}  & \textbf{31.46}  & \textbf{27.91}        \\
                                   & SSIM               & 0.86             & \textbf{0.91}  & \textbf{0.89}  & \textbf{0.80}  & \textbf{0.75}  & \textbf{0.79}  & \textbf{0.87}  & \textbf{0.93}   & \textbf{0.92}   & \textbf{0.86}         \\
                                   & LPIPS              & 0.29             & \textbf{0.11}  & \textbf{0.19}  & \textbf{0.21}  & \textbf{0.39}  & \textbf{0.28}  & \textbf{0.11}  & \textbf{0.11}   & \textbf{0.12}   & \textbf{0.20}         \\
                                   & Model Size (MB)    & \textbf{113.1}   & \underline{176.4}& 115.9        & 137.4          & 133.1          & 416.9          & \underline{534.4}& \underline{672.3}& 831.9        & \underline{347.9}      \\
                                   & Ad-Opt Time (s)    & \textbf{53.74}   & \textbf{66.06} & \textbf{46.45} & \textbf{46.67} & \textbf{58.24} & \textbf{76.81} & \textbf{91.61} & \textbf{105.97} & \textbf{114.55} & \textbf{73.34}   \\ 
                                   & FPS                & \textbf{0.63}    & \textbf{0.27}  & \textbf{0.39}  & \textbf{0.37}  & \textbf{0.31}  & \textbf{0.28}  & \textbf{0.30}  & \textbf{0.37}   & \textbf{0.34}   & \textbf{0.36}     \\ \midrule\midrule
\multirow{7}{*}{\begin{tabular}[c]{@{}c@{}}GS-SLAM\\ (RGB)\end{tabular}} & PSNR (dB) & \textbackslash{} & 17.59   & 18.19   & 16.34   & \textbackslash{}          & 18.25          & 20.01          & 21.48           & 20.57           & 18.92\\
                                   & SSIM               & \textbackslash{} & 0.48           & 0.63           & 0.49           & \textbackslash{}           & 0.66           & 0.48           & 0.65            & 0.66            & 0.58                  \\
                                   & LPIPS              & \textbackslash{} & 0.76           & 0.74           & 0.72           & \textbackslash{}           & 0.62           & 0.58           & 0.68            & 0.75            & 0.69                  \\
                                   & Model Size (MB)    & \textbackslash{} & \textbf{43.6}  & \textbf{22.6}  & \textbf{23.7}  & \textbackslash{}  & \textbf{26.1}  & \textbf{44.6}  & \textbf{38.2}   & \textbf{29.9}   & \textbf{32.67}        \\
                                   & FPS                & \textbackslash{} & \underline{0.19}& \underline{0.19}& \underline{0.37}& \textbackslash{}& \underline{0.26}& \underline{0.23}& \underline{0.21}& \underline{0.26}& \underline{0.24}\\
                                   & Finish Tracking    & \ding{55}        & \Checkmark     & \Checkmark     & \Checkmark     & \ding{55}       & \Checkmark     & \Checkmark     & \Checkmark      & \Checkmark      & \textbackslash{}      \\ \bottomrule

\end{tabular*}%
\vspace{-4pt}
\end{table*}

\vspace{4pt}
\noindent\textbf{Implementation details}. The optimization process of our On-the-Fly GS includes three stages (see Fig. \ref{fig:2}). In phase 1 of initialization, after On-the-Fly SfM completes pose estimation and sparse point cloud generation for the first $N_0=30$ images, we perform 2,000 training iterations on these images. In phase 2, for each newly acquired image, we choose the top $N_m=10$ images with the highest weights as key training targets, and $T_I=200$ iterations of optimization will be performed on the current 3DGS field. Since each new fly-in image triggers 200 iterations of optimization on the current 3DGS field, the average number of training iterations per image will approach 200 as the total number of images increases. Therefore, the parameter $T_a$ is set to 200. Additionally, the preset parameter $\eta^{0}$ is $1.6e-4$ and $\eta^{f}$ is $1.6e-6$. The weights of Load-Balancing-Loss $\lambda_\text{load}$, L1-Loss $\lambda_1$, and SSIM-Loss $\lambda_\text{ssim}$ are set to 0.41, 0.47, and 0.12, respectively. All other settings are consistent with the original 3DGS configuration.


\subsection{Comparisons with other SOTA methods}

We compare On-the-Fly GS with two efficient offline-training solutions (Adr-Gaussian\cite{5} and Taming-3DGS\cite{9}) and two popular SLAM-based 3DGS methods (GS-SLAM\cite{21} and RTG-SLAM\cite{20}). Since RTG-SLAM only accepts RGB-D input, comparisons on RGB datasets are conducted solely among the offline training solutions and GS-SLAM. RTG-SLAM is included in the comparison only when using RGB-D datasets. Unless otherwise specified, all offline training solutions and SLAM-based solutions are evaluated using their default parameter settings.
`
\begin{figure}[htp]
    \centering
    \includegraphics[trim={0 0 0 0},clip, width=\linewidth]{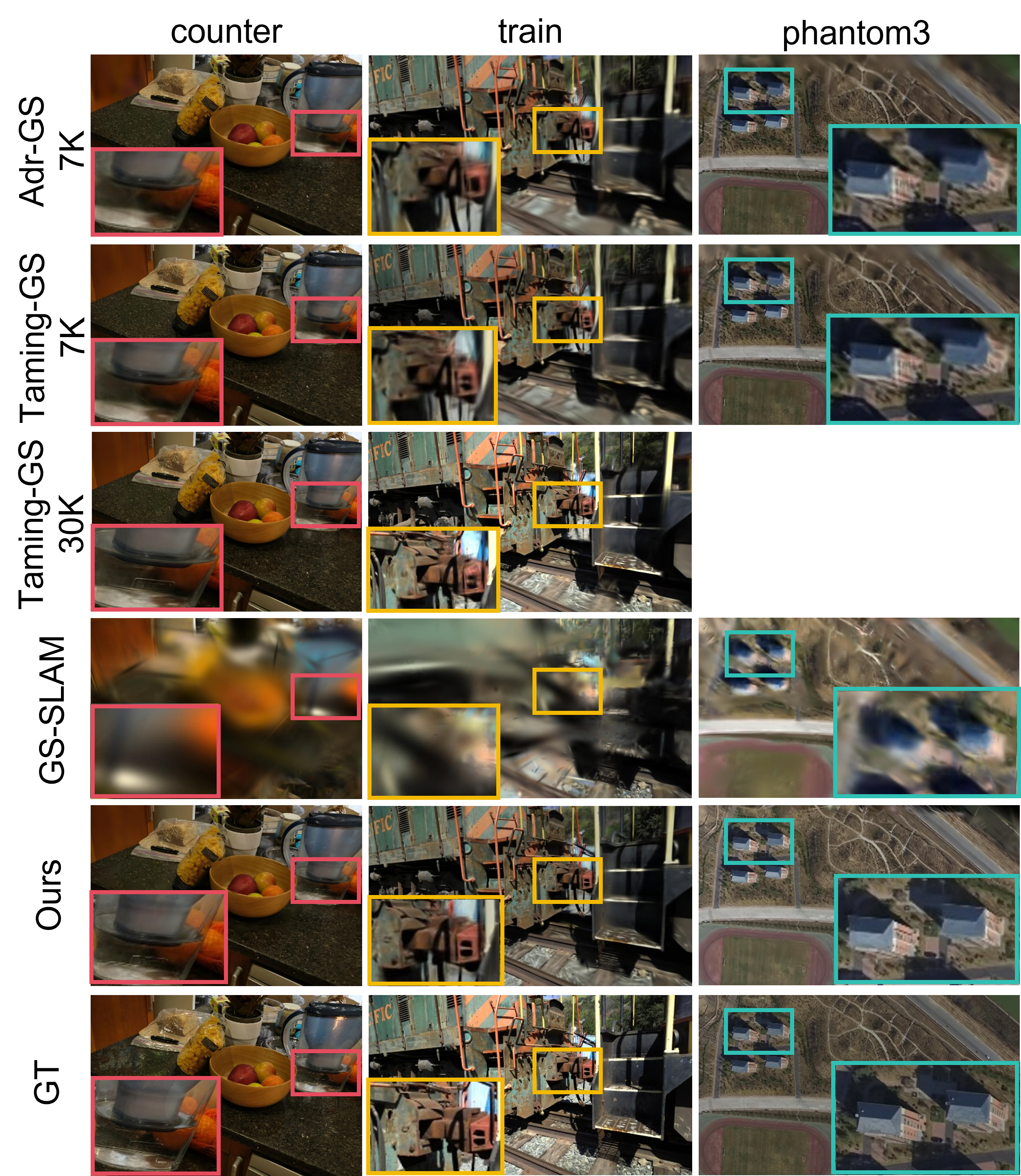}
    \caption{Rendering Results of On-the-Fly GS and other solutions with default settings}
    \label{fig:5_2}
    \vspace{-0.4cm}
\end{figure}

\textbf{Results on RGB Datasets.} To conduct a fair evaluation, the offline-training solutions also use the SfM results from On-the-Fly SfM, but after all images are solved. The time consumption for online SfM is reported in Table \ref{Comparison0}. As On-the-Fly GS and the offline-training solutions undergo the same pose estimation process, we only record the cost time from the completion of pose estimation to the completion of 3DGS field optimization (Ad-Opt Time). Since On-the-Fly GS performs training concurrently with the online SfM, in fact, only the cost time for the additional optimization in Phase 3 is recorded. In contrast, the offline solutions report their total execution time. For GS-SLAM, we report its FPS using the default setting. In our On-the-Fly GS, the reported FPS takes the entire duration of both online SfM and 3DGS field optimization into account. Moreover, since the degree of overlap between image pairs in the RGB datasets is significantly lower than that of typical SLAM datasets, each image is regarded as a keyframe when training GS-SLAM. Table \ref{Comparison0} and Fig. \ref{fig:5_2} show the comparison results on RGB datasets.

\begin{table*}[]
\centering
\caption{Comparison on RGB-D datasets. Best and second-best are highlighted in bold and underlined.}
\vspace{-0.2cm}
\label{Comparison1}

\begin{tabular*}{0.92\linewidth}{cccccccccccc}
\toprule

Method                         & Modality               & Metric      & room0 & room1 & room2 & office0 & office1 & office2 & office3 & office4 & Avg. \\ \midrule
\multirow{5}{*}{GS-SLAM}       & \multirow{5}{*}{RGB}   & PSNR (dB)   &\underline{29.31}  &27.58  &29.24  &33.88    &35.51    &25.24    &28.79    &27.89    &29.89 \\
                               &                        & SSIM        &0.89   &0.84   &0.89   &0.93     &0.94     &0.88     &0.89     &0.91     &0.89  \\
                               &                        & LPIPS       &\underline{0.24}   &0.32   &0.23   &0.18     &0.16     &0.28     &0.18     &0.24     &0.22  \\
                               &                        & FPS         &0.89   &0.92   &0.93   &0.95     &1.05     &0.89     &0.93     &0.91     &0.93  \\ 
                               &                        & ATE-MSE (m) &0.0849 &0.4468 &0.1897 &0.2014   &0.2144   &0.5454   &0.1664   &0.8466   &0.1869\\ \midrule
\multirow{5}{*}{RTG-SLAM}      & \multirow{5}{*}{RGB-D} & PSNR (dB)   &26.45  &\underline{30.04}  &\underline{31.31}  &\underline{36.82}    &\underline{37.37}    &\underline{30.05}    &\underline{30.18}    &\underline{33.28}    &\underline{31.94} \\
                               &                        & SSIM        &\underline{0.92}   &\underline{0.94}   &\underline{0.95}   &\textbf{0.98}     &\textbf{0.99}     &\textbf{0.96}     &\textbf{0.97}     &\textbf{0.97}     &\textbf{0.96}  \\
                               &                        & LPIPS       &0.28   &\underline{0.22}   &\underline{0.19}   &\underline{0.10}     &\textbf{0.11}     &\underline{0.19}     &\underline{0.18}     &\underline{0.16}     &\underline{0.18}  \\
                               &                        & FPS         &\textbf{2.71}   &\textbf{2.94}   &\textbf{2.73}   &\underline{2.71}     &\underline{3.05}     &\textbf{3.13}     &\textbf{2.94}     &\textbf{2.94}     &\textbf{2.89}  \\ 
                               &                        & ATE-MSE (m) &\textbf{0.0019} &\textbf{0.0019} &\textbf{0.0011} &\textbf{0.0015}   &\textbf{0.0013}   &\textbf{0.0024}   &\textbf{0.0023}   &\textbf{0.0025}   &\textbf{0.0019}\\ \midrule
\multirow{5}{*}{Ours}          & \multirow{5}{*}{RGB}   & PSNR (dB)   &\textbf{31.79}  &\textbf{33.67}  &\textbf{34.65}  &\textbf{39.44}    &\textbf{38.77}    &\textbf{34.03}    &\textbf{35.38}    &\textbf{36.65}    &\textbf{35.55} \\
                               &                        & SSIM        &\textbf{0.94}   &\textbf{0.95}   &\textbf{0.96}   &\underline{0.97}     &\underline{0.96}     &\underline{0.95}     &\underline{0.96}     &\underline{0.96}     &\textbf{0.96}  \\
                               &                        & LPIPS       &\textbf{0.12}   &\textbf{0.11}   &\textbf{0.11}   &\textbf{0.09}     &\underline{0.14}     &\textbf{0.13}     &\textbf{0.09}     &\textbf{0.10}     &\textbf{0.11}  \\
                               &                        & FPS         &\underline{2.01}   &\underline{2.23}   &\underline{1.99}   &\textbf{3.25}     &\textbf{3.56}     &\underline{1.86}     &\underline{1.88}     &\underline{2.12}     &\underline{2.36}  \\ 
                               &                        & ATE-MSE (m) &\underline{0.0035} &\underline{0.0054} &\underline{0.0025} &\underline{0.0057}   &\underline{0.0053}   &\underline{0.0070}   &\underline{0.0024} &\underline{0.0075} &\underline{0.0049}\\ \bottomrule

\end{tabular*}%
\vspace{-4pt}
\end{table*}

Compared to the offline solutions with 7K iterations, our On-the-Fly GS achieves significantly higher rendering quality as well as explicitly less cost time. This indicates that, thanks to the proposed Local \& Semi-Global training strategy, our On-the-Fly GS can quickly achieve high-quality rendering with just a small number of additional iterations. Furthermore, On-the-Fly GS exhibits a pronounced advantage on the UAV dataset, significantly outperforming other solutions in both training efficiency and rendering quality. These results underscore the strong potential of the proposed approach for large-scale UAV image dataset applications. With 30K training iterations, Taming-3DGS is unable to complete training on the UAV image datasets due to excessive memory consumption. As for other scenes, On-the-Fly GS consistently achieves over 80\% of the rendering quality delivered by Taming-3DGS, while exhibiting clear advantages in terms of model size and cost time.

Due to the significantly low overlapping degree between consecutive image pairs in the datasets presented in Table \ref{Comparison0}, GS-SLAM exhibits extremely poor camera tracking performance across all the scenes, and even fails to track on some datasets (skfx and Temple). Additionally, the tested RGB datasets have high resolutions and very complex scene content, which leads to an increase in the training time for GS-SLAM and a substantially lower rendering quality compared to other methods. These results reveal the potential limitations of SLAM-based 3DGS methods on handling datasets without a reliable depth map or containing consecutive image pairs of low overlapping degree.

\textbf{Results on RGB-D Datasets.} To compare the performance of SLAM-based 3DGS method, the Replica dataset with depth priors is tested on the two popular pipelines - GS-SLAM\cite{21} and RTG-SLAM\cite{20}. Since these two methods' solutions perform 3DGS field optimization using only the selected keyframes, we retain the selected keyframes from GS-SLAM, which are subsequently used as input for our On-the-Fly GS.

Table \ref{Comparison1} shows the comparison results on the RGB-D dataset. It can be seen that, despite only RGB information from keyframes is used as input, the proposed On-the-Fly GS achieves the best or the second-best rendering quality among all evaluated methods and datasets. Although On-the-Fly GS is slightly inferior to RTG-SLAM in terms of time efficiency and tracking accuracy, our method attains superior rendering quality without requiring depth and does not rely on sequential video frames as input. Instead, it only requires that newly added images have a certain degree of overlap with any previously registered views. These characteristics make On-the-Fly GS more broadly applicable than existing SLAM-based solutions.

\subsection{Ablation Studies}

\begin{table}[]
\caption{Ablation experiment of the proposal components in the On-the-Fly GS method.}
\vspace{-6pt}
\label{Ablation}
\resizebox{\columnwidth}{!}{%
\begin{tabular}{cccccc}
\toprule
Methods  & Ad-Opt Time (s) & Model Size (MB)   & PSNR  & SSIM & LPIPS \\ \midrule
No\_FU   &\underline{110.31}&\textbf{541.1}    &30.35  &0.88  &0.21   \\
No\_IW   &116.76           &\underline{575.7}  &31.99  &0.91  &0.15   \\
No\_GO   &161.83           &860.3              &32.17  &0.92  &0.13   \\
No\_Load &252.04           &1315.4             &\textbf{34.13}&\textbf{0.95}&\textbf{0.09}\\
No\_SPB  &213.59           &674.5              &33.14  &0.93  &0.11   \\
Ours     &\textbf{105.97}  &672.3              &\underline{33.29}&\underline{0.93}&\underline{0.11}\\ \bottomrule
\end{tabular}%
}
\vspace{-0.6cm}
\end{table}

To explore the efficacy of each component proposed in this work, an extensive ablation study is conducted. Based on the \textit{npu} dataset, we first turn on all the proposed component (as Ours) and then sequentially switch off one of the following strategies: 1) 3DGS Field Update (No\_FU). 2) Image Weighting and Training Iteration Allocation (No\_IW). 3) Semi-Global Optimization (No\_GO): only train new fly-in image and its neighbors for $N_I=200$ iterations, and no training for the very previous images. 4) Load Balancing (No\_Load). 5) Splatting-based Parallelism in Backpropagation (No\_SPB). 

The corresponding quantitative results are presented in Table \ref{Ablation}. We can see that, removing any of the proposed components can negatively impact the 3DGS training, proving their corresponding effectiveness: 1) Disabling 3DGS Field Update prevents the addition of new Gaussians during scene expansion, resulting in few Gaussians allocated in the newly observed region. 2) Image Weighting and Training Iteration Allocation guide the training process to focus more on the newly expanded region. The optimization of the newly observed region proceeds slowly without this strategy. 3) Removing Global Optimization makes the 3DGS training over-emphasize the optimization on newly observed regions, leading to an excessive accumulation of Gaussians while neglecting rendering quality in previous regions. 4) Load Balancing optimizes the spatial distribution of Gaussians and removes redundant ones, effectively reducing model size while improving training efficiency. 5) Splatting-based Parallelism in Backpropagation accelerates the optimization of the 3DGS field by significantly accelerating the backpropagation operation.



\section{CONCLUSION}

In this paper, we propose On-the-Fly GS, a novel near real-time 3DGS training framework via a progressive optimization manner, which can dynamically output 3DGS field during image capturing. First, we introduce an self-adaptive method for allocating different training iterations to each newly captured image and its neighboring images during 3DGS optimization, enabling fast training for the newly expanded region. Second, we present a new learning rate updating solution based on image rendering quality, adaptively estimating the learning rate for each training iteration based on the training progress of the current image and its neighboring connected images. Finally, inspired by Adr-Gaussian and Taming 3DGS, we further control the number of Gaussians and improve the training efficiency. The experimental results demonstrate that our On-the-Fly GS can successfully achieve near real-time 3DGS training and considerable rendering performance, even when the input images lack temporal and spatial continuity or are not accompanied by depth. Compared to existing SLAM-based 3DGS methods, On-the-Fly GS exhibits greater versatility and is feasible for a wider range of practical applications.


\bibliographystyle{IEEEtran.bst}
\bibliography{IEEEabrv, references}

\end{document}